\documentclass[runningheads]{llncs}
\usepackage{graphicx}
\usepackage{xcolor}
\usepackage{amssymb}
\usepackage{pifont}
\newcommand{\xmark}{\ding{55}}
\newcommand\blfootnote[1]{%
  \begingroup
  \renewcommand\thefootnote{}\footnote{#1}%
  \addtocounter{footnote}{-1}%
  \endgroup
}
\usepackage[hyperfootnotes=false]{hyperref}
\begin{document}
\title{FLex: Joint Pose and Dynamic Radiance Fields Optimization for Stereo Endoscopic Videos}

\titlerunning{FLex: Joint Pose and Dynamic RF Optimization for Endoscopic Videos}
%
\author{Florian Philipp Stilz\inst{*,1,2} \and
Mert Asim Karaoglu\inst{*,1,2} \and
Felix Tristram\inst{*,1} \and 
Nassir Navab \inst{1} \and
Benjamin Busam\inst{1} \and
Alexander Ladikos\inst{2}
}

%
\authorrunning{F. Stilz et al.}
%

\institute{Technical University Munich \and
ImFusion GmbH
}
\maketitle              
\begin{abstract}

Reconstruction of endoscopic scenes is an important asset for various medical applications, from post-surgery analysis to educational training.
Neural rendering has recently shown promising results in endoscopic reconstruction with deforming tissue.
However, the setup has been restricted to a static endoscope, limited deformation, or required an external tracking device to retrieve camera pose information of the endoscopic camera.
With FLex we adress the challenging setup of a moving endoscope within a highly dynamic environment of deforming tissue.
We propose an implicit scene separation into multiple overlapping 4D neural radiance fields (NeRFs) and a progressive optimization scheme jointly optimizing for reconstruction and camera poses from scratch.
This improves the ease-of-use and allows to scale reconstruction capabilities in time to process surgical videos of 5,000 frames and more; an improvement of more than ten times compared to the state of the art while being agnostic to external tracking information.
Extensive evaluations on the StereoMIS dataset show that FLex significantly improves the quality of novel view synthesis while maintaining competitive pose accuracy.

\keywords{3D Reconstruction  \and Neural Rendering \and Robotic Surgery}
\end{abstract}

\blfootnote{$^{*}$ The authors contributed equally. \\Contact author: Florian Philipp Stilz (\textit{florian.stilz@tum.de}).}
\section{Introduction}
\label{sec:intro}

\begin{figure*}[t]
    \centering
    \includegraphics[width=1.0\textwidth]{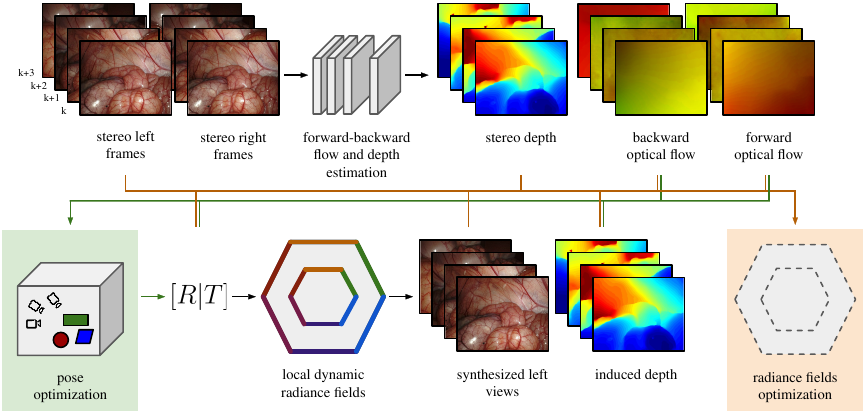}
    \caption{Joint pose and radiance fields optimization. $k$ indexes the frames along the temporal dimension. Orange and green arrows show the flow of inputs and outputs in the optimization processes of pose and radiance fields.}
    \label{fig:architecture}
\end{figure*}
Visually and geometrically accurate reconstructions of surgical scenes are crucial components for various computer vision and AR/VR applications such as post surgical longitudinal assessment~\cite{liu2020reconstructing}, surgical training~\cite{lange2000virtual}, as well as for data generation for other learning-based computer vision and robotics applications~\cite{long2022integrating}.
However, endoscopic videos present a range of visual and practical difficulties for contemporary reconstruction methods, including strong non-homeomorphic deformations, prolonged recording times and the challenge of determining camera positions.
Often, this leads to a dependency on external tools for acquisition, diminishing the ease-of-use of the reconstruction framework.

Prior methods widely explore usage of explicit representations such as sparse and dense point clouds~\cite{song2017dynamic,rodriguez2023nr} in visual odometry and simultaneous localization and mapping (SLAM) frameworks.
Even though these approaches often provide efficient solutions for combined camera tracking and reconstruction, their incomplete geometry modeling results in limitations when rendering views from new camera poses.
Our method is similar regarding joint optimization of localization and reconstruction, however, we instead use a dynamic neural radiance fields (NeRF)~\cite{mildenhall2021nerf} based architecture to reconstruct the scene together with capturing the dynamics, thereby enabling high quality time dependent novel view synthesis.

EndoNeRF \cite{wang2022endonerf} is the first in the line of works that adapt a dynamic NeRF archicture~\cite{pumarola2021d} for endoscopic scenes.
EndoSurf \cite{zha2023endosurf} builds on top of the previous work and substitutes the representation from volumetric density to a signed-distance function (SDF)~\cite{wang2021neus}.
LerPlane~\cite{yang2023lerplane} and its follow-up work ForPlane~\cite{yang2023efficient} utilize explicit data structures as in \cite{chen2022tensorf,kplanes,cao2023hexplane} for faster optimization and higher rendering quality. 
While these works present great results and a promising research direction, unlike our method, they rely on external, reliable measurement or computation of the camera poses which are difficult to obtain in endoscopic environments.
While EndoNeRF~\cite{wang2022endonerf}, EndoSurf~\cite{zha2023endosurf}, and LerPlane~\cite{yang2023lerplane} only test their methods on the EndoNeRF dataset~\cite{wang2022endonerf}, which consists of a static camera and has around 150 frames per sequence, ForPlane~\cite{yang2023efficient} also tests on the Hamlyn dataset~\cite{stoyanov2005soft,mountney2010three}, containing around 301 frames per sequence, utilizing the camera poses estimated using Endo-Depth-and-Motion~\cite{recasens2021endo}.
In our experiments, we extend these investigations to recordings with moving cameras along with tissue deformations with significantly longer durations of up to 5,000 frames.

More recently, various works investigated pose optimization within NeRF setups integrating core components of visual odometry (VO) and SLAM pipelines, with experiments on natural scenes.
NeRF-\phantom{}-~\cite{wang2021nerf} and BARF~\cite{lin2021barf}, propose solutions for jointly learning the poses as a part of NeRF optimization.
LocalRF~\cite{meuleman2023localrf} further extends this idea for larger scale scenes.
However, unlike in the case of endoscopic scenes, these methods employ radiance fields with static scene assumptions, and, as we show in our experiments, tend to have a performance drop when confronted with highly dynamic content.

In this work, we present FLex; a NeRF-based architecture capable of high quality novel view synthesis from pose-free surgical videos, containing strong deformations.
Employing a progressive optimization scheme~\cite{meuleman2023localrf}, FLex utilizes local dynamic radiance fields to jointly optimize for pose and scene representation.
In the context of NeRF applications on endoscopic scenes, along with a concurrent work following a different approach, BASED~\cite{saha2023based}, we believe FLex is the first to investigate joint pose optimization.
In addition to architectural differences, in this work we jointly target efficient scaling to longer sequences.
Extensively evaluated on the StereoMIS~\cite{hayoz2023learning} dataset, our method shows state-of-the-art results in novel view synthesis.
Furthermore, it achieves competitive tracking accuracy compared to recent methods designed specifically for this task~\cite{hayoz2023learning}.

To summarize our contributions are the following: 
\begin{itemize}
    \item We present a novel NeRF architecture for the challenging task of 4D reconstruction in highly dynamic endoscopic scenes in the absence of camera pose information, achieved by joint optimization in a progressive strategy.
    \item Dissecting the scene into multiple smaller 4D models with overlaps allows to efficiently scale reconstruction in time. We conduct experiments to showcase this on up to 5,000 frames, more than ten times the size used in previous works. Together with our first contribution this improves the applicability to real world use-cases.
    \item Experimental results on the StereoMIS dataset reveal a clear advantage in terms of novel view synthesis with competitive accuracy in camera poses.
\end{itemize}
\section{Method}
\label{sec:method}

\subsection{Overview}

\begin{figure*}[t]
    \centering
    \includegraphics[width=0.6\textwidth]{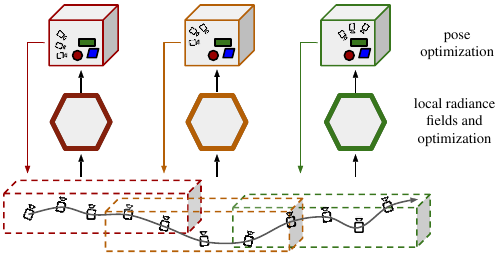}
    \caption{Joint progressive pose and local dynamic radiance fields optimization. Spatial extents clustered within the bounding boxes of different colors represent the spatio-temporal domain of the corresponding local radiance fields. The arrow on the camera trajectory shows the temporal direction.}
    \label{fig:progressive_optimization}
\end{figure*}
Given a rectified stereo-endoscopic video, our goal is to reconstruct the 4D scene accurately without prior camera pose information.
For this we propose a new method \textbf{FLex}, standing for \textbf{F}low-optimized \textbf{L}ocal H\textbf{ex}planes, depicted in Fig. \ref{fig:architecture}, and combines advancements from recent NeRF literature to build multiple smaller dynamic models that are progressively optimized. In contrast to prior work \cite{zha2023endosurf,wang2022endonerf,yang2023efficient}, we do not have one unified representation of the scene but multiple smaller overlapping ones. The representation of local models allows us to represent larger scenes, both temporally and spatially, accurately without incurring prohibitive memory growth while maintaining a high feature grid resolution. Furthermore, adopting a progressive optimization scheme enables the optimization of poses from scratch. For further regularization, we add supervision through optical flow and stereo depth.

\subsection{4D Scene Representation}
NeRFs\cite{mildenhall2021nerf} implicitly model a 3D scene utilizing differentiable volume rendering to predict pixel colors.
They can be adapted to a 4D scene representation by adding the timestep $k$ as an additional input to the model.
HexPlane \cite{cao2023hexplane} models a dynamic scene using an explicit 4D feature grid paired with an implicit MLP. The grid is constructed from several planes, where each plane represents a combination of two dimensions, yielding six planes in total. During ray-casting, the corresponding features on each plane are extracted for the spatio-temporal locations and combined by multiplication and concatenation before being fed into smaller MLPs and similarly to NeRF \cite{mildenhall2021nerf} rendered with volumetric rendering. 

\subsection{Progressive Optimization}
Endoscopic videos contain two main challenges for NeRF architectures: They are dependent on external tools for accurate pose estimation and can constitute arbitrarily long sequences.
In order to tackle these two problems in a robust and efficient way, our joint pose and radiance fields optimization scheme utilizes the concepts of progressive optimization and dynamic allocation of local HexPlane models as visualized in Fig. \ref{fig:progressive_optimization} and inspired by LocalRF~\cite{meuleman2023localrf}.

In the scope of progressive optimization, we start with the first five frames of the sequence, then we consecutively add one frame at a time, initializing it's camera pose parameters with the prior camera pose.
When the appended frame increases the number of frames above a preset threshold, $t_{k}$, or the distance between the optimized position of the camera and the center of the current local model is larger than a distance threshold, $t_d$, we instantiate a new local model, setting this new frame to be its origin.
To ensure consistency across local models, we assign the last thirty frames of the previous model to be overlapping with the new local model.
To secure a coherent trajectory during progressive optimization, we consistently sample rays from the last four appended frames.
When a new local model is initiated, the weights of the previous one are frozen and offloaded from the GPU to prevent unnecessary memory usage.
During inference, if a pose corresponds to the spatial and temporal extent of multiple local models, each model's contribution is aggregated into the ray-casting formulation with blending weights linearly set on the overlapping regions based on the proximity to the centers of the local models.
Before a new local model is initialized, the last model goes into a refinement phase where the pose and model parameters are optimized with batches of samples uniformly picked along its entire span.

\begin{figure*}[t]
    \centering
    \includegraphics[width=0.85\textwidth]{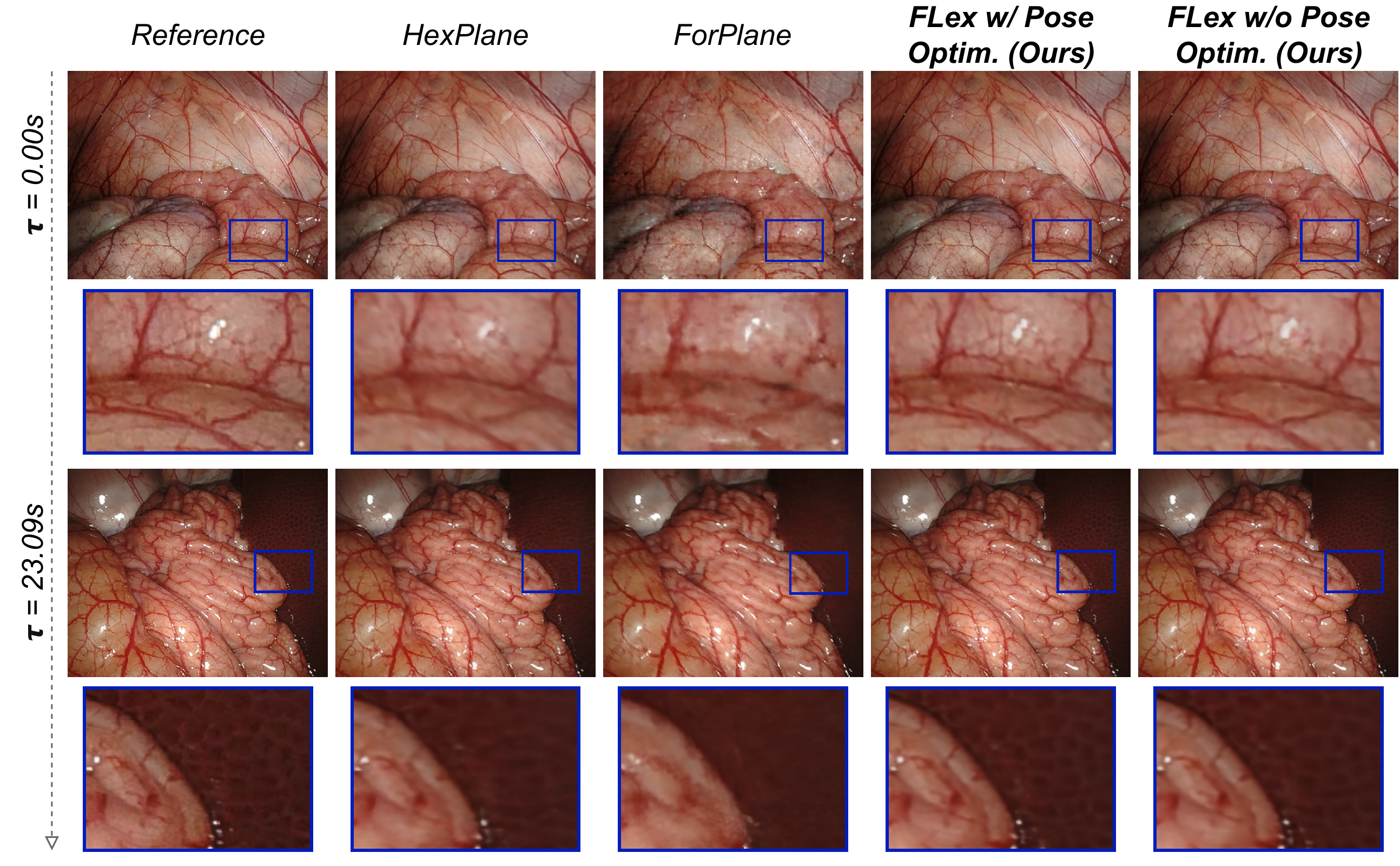}
    \caption{Qualitative results on a 1,000 frame scene with breathing deformations and camera motion.}
    \label{fig:qual_results}
\end{figure*}
\subsection{Training Objectives}
We employ the common photometric loss $\mathcal{L}_{rgb}$ as defined in Eq. (\ref{E: rgb}) with ground-truth $C(\mathbf{r})$ and predicted $\hat{C}(\mathbf{r})$ pixel values for ray $\mathbf{r}$ within the set of rays $\mathcal{R}$:
\begin{equation}
    \label{E: rgb}
    \mathcal{L}_{rgb} = \frac{1}{|\mathcal{R}|}\sum_{\mathbf{r}\in\mathcal{R}}\left\|\hat{C}(\mathbf{r}) - C(\mathbf{r})\right\|_{2}^{2}
\end{equation}

Additionally, we also use the depth supervision loss along with the line-of-sight prior as introduced by Rematas et al.~\cite{rematas2022urban}.
We denote them together as $\mathcal{L}_{z}$.
The line-of-sight prior regularizes the density values along a ray to be concentrated on the actual surface, thereby, together with the depth loss, improving the capture of the scene geometry.

Furthermore, our method is optimized via an optical flow loss $\mathcal{L}_{f}$ in both temporal directions as described in Eq. (\ref{E: Optical_Flow_Loss}). The estimated optical flow $\hat{F}_{k\rightarrow k\pm1}$ is induced via finding the surface point for a given ray at time $k$ in 3D with the help of the predicted depth using the projection from 2D to 3D $\pi_{3D}$ and then transforming the point to the adjacent timestep $k\pm1$ using the relative camera extrinsics $\left[ R|T\right]_{k\rightarrow k\pm1}$. The resulting 3D point is then projected from 3D to 2D via $\pi_{2D}$ using the known camera intrinsics and compared to the initial pixel location $\mathbf{p}(\mathbf{r})$ at time k, see Eq. (\ref{E: Optical_Flow}). 
\begin{equation}
    \label{E: Optical_Flow}
    \hat{\mathcal{F}}_{k\rightarrow k\pm1}(\mathbf{r}) = \mathbf{p}(\mathbf{r}) - \pi_{2D} \left(\left[R|T\right]_{k\rightarrow k\pm1} \pi_{3D}(\mathbf{r}, \hat{D}) \right)
\end{equation}
\begin{equation}
    \label{E: Optical_Flow_Loss}
    \mathcal{L}_{f} = \frac{1}{|\mathcal{R}|}\sum_{\mathbf{r}\in\mathcal{R}}
    \left\|\hat{\mathcal{F}}_{k\rightarrow k\pm1}(\mathbf{r}) - \mathcal{F}_{k\rightarrow k\pm1}(\mathbf{r})\right\|_{1}
\end{equation}
All the aforementioned losses are aggregated into our final loss function $\mathcal{L}$. It is essential to highlight that we employ all three loss terms to optimize our method FLex. The camera extrinsics are only optimized by the optical flow loss $\mathcal{L}_{f}$.

Note that the optical flow loss \(\mathcal{L}_{f}\) is entirely removed after 20\% of the refinement iterations to ensure an early pose convergence.
\begin{equation}
    \label{E: Final_Loss}
    \mathcal{L} = \mathcal{L}_{rgb} + \lambda_z \mathcal{L}_{z} +  \lambda_f \mathcal{L}_{f}
\end{equation}
\section{Experiments}
\label{sec:exp}
\subsection{Dataset and Evaluation Metrics}
We systematically assess the efficacy of our approach using the publicly available StereoMIS \cite{hayoz2023learning} dataset, recorded using a stereo endoscope of a da Vinci Xi robot; ground-truth camera trajectories are measured using the forward kinematics.
In total we extract five sequences for general comparison, each 1,000 frames long equivalent to ca. 29 seconds per clip.
Two of these scenes exhibit concurrent breathing motion and camera movement, while another two showcase pronounced non-rigid deformations induced by surgical tools amidst subtle changes in camera perspective.
The remaining scene presents a nearly static environment accompanied by rapid camera movements.
Furthermore, we create two additional longer sequences to study the method's behavior given a larger temporal and spatial extent.
One sequence contains deformations induced by surgical tools and consists of 5,000 frames, while the other scene incorporates more rapid camera motion and comprises 4,000 frames.
For measuring our method quantitatively, we follow the evaluation of \cite{wang2022endonerf} by making use of PSNR, SSIM, and LPIPS metrics (denoted with subscripts "a" and "v" for AlexNet~\cite{krizhevsky2012imagenet} and VGG~\cite{simonyan2014very} backbones) and L1-distance metric, indicated in mm, to assess the captured geometry by comparing the induced and stereo-estimated depth images.
We evaluate the estimated camera poses using root-mean-squared absolute trajectory error (ATE-RMSE), relative translational and rotational pose errors (RPE-Trans and RPE-Rot).
\subsection{Implementation Details}
In our local models, we set the dimension of the spatial feature grids to 512 for \((x,y,z)\), and the temporal dimension is set to half of the image sequence length of a scene.
The feature dimension is 72 in total for both density and color.
For a fair comparison, we ensure equal capacity for all methods using explicit data structures \cite{cao2023hexplane,meuleman2023localrf,yang2023efficient}. 
Additionally, we adopt a coarse-to-fine approach as in HexPlane \cite{cao2023hexplane} to start with a lower grid resolution and increase over time to the settings mentioned above.
Except for our method with pose optimization and LocalRF~\cite{meuleman2023localrf}, we use Robust-Pose Estimation~\cite{hayoz2023learning} to estimate the camera poses.
In the experiment tables, we use HexPlane\(^{\dagger_1}\) to denote an improved version with scene contraction, depth loss, and optical flow loss; and LocalRF\(^{\dagger_2}\) to depict a version of it which substitutes the monocular with the stereo depth estimation.

Furthermore, we set $\lambda_{z} = 0.01$ and $\lambda_{f} = 1.0$ as illustrated in Eq. (\ref{E: Final_Loss}) and $t_{k}=100$ and $t_{d} = 1.0$.
Overall, we train our method without pose optimization for 100 iterations per frame with a batch size of 4,096 rays, which takes approximately 7 hours on up to 40 GB of an Nvidia Tesla A100 and FLex with pose optimization for an additional 100 iterations per frame during the prior progressive optimization which yields ca. 20 hours on the same hardware configuration.
In comparison, HexPlane, for the exact same settings, takes ca. 6 hours to train. 
We do not mask the tools in any experiment.
In addition, we use RAFT \cite{RAFT} for estimating both optical flow from frame-to-frame ${\mathcal{F}}_{k\rightarrow k\pm1}(\mathbf{r})$ for both directions and for obtaining stereo depth $D$, which we both use as pseudo-ground-truth for model optimization.
\begin{table}[t]
\centering
\caption{View synthesis quality on StereoMIS dataset. The metrics are computed as an average for five 1,000 frame sequences. L1-Distance is computed between the synthesized and the ground truth depth images in mm. The best result for each metric is marked in bold.}
\label{Tab: Avg_Results}
\scalebox{0.9}{
\begin{tabular}{l|l|l|l|l|l}
\hline
Model    & PSNR $\uparrow$ & SSIM $\uparrow$ & LPIPS$_a$ $\downarrow$ & LPIPS$_v$ $\downarrow$ & L1-Distance $\downarrow$ \\ \hline
EndoNeRF~\cite{wang2022endonerf} & $21.99$    & $0.590$    & $0.496$        & $0.514$  & $-$        \\
EndoSurf~\cite{zha2023endosurf} & $25.18$    & $0.622$    & $0.528$        & $0.529$  & $8.105$       \\
ForPlane~\cite{yang2023efficient} & $30.35$    & $0.783$    & $0.208$        & $0.301$  & $23.717$       \\
LocalRF\(^{\dagger_2}\)~\cite{meuleman2023localrf}  & $27.41$    & $0.781$    & $0.245$        & $0.288$ & $4.576$        \\
HexPlane\(^{\dagger_1}\)~\cite{cao2023hexplane} & $30.85$    & $0.819$    & $0.211$        & $0.273$  & $1.532$      \\
\hline
FLex w/o Pose Optim. (Ours)     & $\mathbf{31.10}$    & $\mathbf{0.836}$    & $0.200$        & $\mathbf{0.244}$  & $1.456$        \\
FLex w/ Pose Optim. (Ours)    & $30.62$    & $0.818$    & $\mathbf{0.179}$        & $0.245$   & $\mathbf{1.273}$     \\ \hline 
\end{tabular}}
\end{table}
\subsection{Quantitative and Qualitative Results}
We conduct a comprehensive comparison of our method, both with and without pose optimization, against the latest state-of-the-art (SoTA) NeRF methods designed for endoscopy \cite{yang2023efficient,wang2022endonerf,zha2023endosurf} and two additional baselines \cite{cao2023hexplane,meuleman2023localrf}. The results in Table \ref{Tab: Avg_Results}, summarizing the average results across all 5 scenes, demonstrate that our method without pose optimization consistently outperforms all baselines and notably surpasses the current endoscopic SoTA, ForPlane. In addition our method with pose optimization (FLex w/ Pose Optim.) also manages to outperform ForPlane in all metrics and is competitve to ours without pose optimization (FLex w/o Pose Optim.) and the improved HexPlane. These quantitative findings are substantiated by our qualitative results presented in Fig. \ref{fig:qual_results}, highlighting that FLex, with and without prior poses, achieves superior image quality compared to the most competitive baselines \cite{cao2023hexplane,yang2023efficient}.
\subsection{Impact of Sequence Length on View Synthesis Quality}
We evaluate our model's effect on longer sequences.
Ensuring a reasonable hardware memory cap of 16 GB vRAM, we set the spatial grid sizes to 128. Additionally, we set the maximum temporal dimensions for our method and HexPlane to 50 per local model and 100, respectively, and we train the models for 100 iterations per frame.
As displayed in Table~\ref{Tab: Large_Scene_Results}, our method preserves its reconstruction quality scaling to the longer sequences and achieves a higher performance difference to HexPlane in comparison to the difference on shorter sequences.
\subsection{Pose Accuracy}
We compare our method against a SoTA method in visual odometry for endoscopic scenes, Robust-Pose Estimation~\cite{hayoz2023learning}, on 3 sequences each with 1,000 frames.
As highlighted in Table~\ref{Tab: Avg_Pose_Results}, FLex performs competitively achieving close results to the baseline.
However, we emphasize that this task is not the main focus of our work and can be improved using robust optimization and globally consistent methods in the future. 
\begin{table}[t]
\centering
\caption{Ablation study on longer StereoMIS sequences. L1-distance is computed between the synthesized and the estimated stereo depth images in mm. The best results  are marked in bold.}
\label{Tab: Large_Scene_Results}
\scalebox{0.9}{
\begin{tabular}{l|l|l|l|l|l|l}
\hline
Model  & Frame \#  & PSNR $\uparrow$ & SSIM $\uparrow$ & LPIPS$_a$ $\downarrow$ & LPIPS$_v$ $\downarrow$ & L1-Distance $\downarrow$ \\ \hline
HexPlane\(^{\dagger_1}\)~\cite{cao2023hexplane} & 4,000 & $24.79$    & $0.614$    & $0.545$        & $0.510$  & $4.856$      \\
FLex w/o Pose Optim. (Ours)  & 4,000   & $\mathbf{26.09}$    & $\mathbf{0.661}$    & $\mathbf{0.498}$        & $\mathbf{0.469}$  & $\mathbf{3.567}$        \\ \hline
HexPlane\(^{\dagger_1}\)~\cite{cao2023hexplane} & 5,000 & $28.55$    & $0.718$    & $0.453$        & $0.471$  & $1.902$      \\
FLex w/o Pose Optim. (Ours)  & 5,000   & $\mathbf{29.97}$    & $\mathbf{0.773}$    & $\mathbf{0.386}$ & $\mathbf{0.413}$  & $\mathbf{1.704}$ \\  \hline 
\end{tabular}}
\end{table}
\begin{table}[]
\centering
\caption{Average Pose accuracy on StereoMIS dataset. ATE-RMSE and RPE-Trans are in mm, RPE-Rot is in degrees. The best results are marked in bold.}
\label{Tab: Avg_Pose_Results}
\scalebox{0.9}{
\begin{tabular}{l|l|l|l}
\hline
Model    & ATE-RMSE $\downarrow$ & RPE-Trans $\downarrow$ & RPE-Rot $\downarrow$ \\ \hline
Robust-Pose Estimation~\cite{hayoz2023learning} & $\mathbf{2.164}\pm2.68e-1$    & 
$\mathbf{0.073}\pm3e-5$    & $\mathbf{0.043}\pm2e-6$       \\
LocalRF\(^{\dagger_2}\)~\cite{meuleman2023localrf} & $7.704\pm1.506$    & $0.160\pm8e-4$    & $0.119\pm2e-5$     \\
\hline
FLex w/ Pose Optim. (Ours)    & $2.565\pm1.6e-1$    & $0.127\pm9e-4$    & 
$0.102\pm4e-6$     \\ \hline
\end{tabular}}
\end{table}
\section{Conclusion}
\label{sec:conclusion}
In this work, we present FLex, a novel method for reconstructing pose-free, long surgical videos with challenging tissue deformations and camera motion.
Our approach successfully eliminates the reliance on prior poses by jointly optimizing for reconstruction and camera trajectory.
FLex improves upon the scalability of dynamic NeRFs for larger scenes thus becoming more applicable to realistic surgical recordings, while improving over current methods on the StereoMIS dataset in terms of novel view synthesis with competitive pose accuracy.
We believe that FLex can pave the way towards more easily accessible, realistic and reliable 4D endoscopy reconstructions to improve post surgical analysis and medical education.

%
%
%
\bibliographystyle{splncs04}
\bibliography{main}


\clearpage
\title{FLex: Joint Pose and Dynamic Radiance Fields Optimization for Stereo Endoscopic Videos - Supplementary Material}
\author{Florian Philipp Stilz\inst{*,1,2} \and
Mert Asim Karaoglu\inst{*,1,2} \and
Felix Tristram\inst{*,1} \and 
Nassir Navab \inst{1} \and
Benjamin Busam\inst{1} \and
Alexander Ladikos\inst{2}
}
\institute{Technical University Munich \and
ImFusion GmbH
}
\authorrunning{F. Stilz et al.}
\maketitle

\begin{table}[]
\centering
\caption{Per sequence comparison of pose accuracy on StereoMIS dataset. LocalRF\(^{\dagger_2}\) is the vanilla LocalRF, but optimized via stereo depth. The ATE-RMSE and the RPE-Trans are in mm, and the RPE-Rot is in degrees.}
\label{Tab: Pose_Results}
\scalebox{0.9}{
\begin{tabular}{l|l|l|l|l|l|l}
\hline
Model   & Deform. & \begin{tabular}[c]{@{}l@{}}Camera \\ Motion\end{tabular} & Tool    & ATE-RMSE $\downarrow$ & RPE-Trans $\downarrow$ & RPE-Rot $\downarrow$ \\
\hline
Robust Pose Estimation & \checkmark & \checkmark & \xmark & $2.407$    & $\mathbf{0.068}$    & $\mathbf{0.043}$ \\
        & \xmark & \checkmark & \xmark & $2.640$    & $\mathbf{0.080}$    & $\mathbf{0.054}$   \\
        & \checkmark & \checkmark & \xmark & $\mathbf{1.444}$        & $\mathbf{0.071}$         & $\mathbf{0.032}$        \\ \cline{2-7} 
        & \multicolumn{3}{l}{Average
        }  & $\mathbf{2.164}\pm2.68e-1$         &  $\mathbf{0.073}\pm3e-5$         &  $\mathbf{0.043}\pm2e-6$\\ \hline
LocalRF\(^{\dagger_2}\) & \checkmark & \checkmark & \xmark & $8.210$    & $0.155$    & $0.136$ \\
        & \xmark & \checkmark & \xmark & $8.888$    & $0.198$    & $0.143$   \\
        & \checkmark & \checkmark & \xmark & $6.013$        & $0.128$         & $0.079$        \\
 \cline{2-7} 
        & \multicolumn{3}{l}{Average
        }  & $7.704\pm1.506$         &  $0.436\pm8e-4$         &  $0.119\pm2e-5$\\ \hline
Ours w/ Pose Optim.   & \checkmark & \checkmark & \xmark & $\mathbf{2.106}$        & $0.099$         & $0.090$        \\
        & \xmark & \checkmark & \xmark & $\mathbf{2.509}$        & $0.113$         & $0.123$        \\
        & \checkmark & \checkmark & \xmark & $3.081$        & $0.168$         & $0.093$        \\\cline{2-7} 
        & \multicolumn{3}{l}{Average
        }  &  $2.565\pm1.6e-1$        & $0.127\pm9e-4$          &  $0.102\pm4e-6$        \\ \hline
\end{tabular}
}
\end{table}

\begin{table}[]
\centering
\caption{Per sequence comparison of novel synthesis quality on StereoMIS dataset. L1 Distance is computed between the synthesized and the ground truth depth images in mm. HexPlane$^{\dagger_1}$ stands for an optimized baseline with scene contraction, depth loss, and optical flow loss, while LocalRF$^{\dagger_2}$ is the vanilla LocalRF, but optimized via stereo depth.\colorbox{blue!20}{Blue} indicates the best result and \colorbox{red!20}{red} the second best for each metric, respectively.}
\label{Tab: All_Results}
\scalebox{0.85}{
\begin{tabular}{l|l|l|l|l|l|l|l|l}
\hline
Model                     & Deform. & \begin{tabular}[c]{@{}l@{}}Camera\\ Motion\end{tabular} & Tool    & PSNR $\uparrow$  & SSIM $\uparrow$   & LPIPS$_a$ $\downarrow$ & LPIPS$_v$ $\downarrow$ & L1 Distance $\downarrow$ \\ \hline
EndoNeRF                  & \checkmark & \checkmark & \xmark & $25.28$ & $0.628$ & $0.507$   & $0.500$ & $-$   \\
                          & \xmark & \checkmark & \xmark & $11.22$     & $0.546$      & $0.517$        & $0.573$ & $-$       \\
                          & \checkmark & \xmark & \checkmark & $23.87$     & $0.569$      & $0.501$        & $0.517$  &$-$      \\
                          & \checkmark & \checkmark & \xmark & $26.56$     & $0.636$      & $0.409$        & $0.456$   &$-$     \\
                          & \checkmark & \xmark & \checkmark & $23.03$     & $0.569$      & $0.545$        & $0.524$  &$-$      \\ \hline
EndoSurf                  & \checkmark & \checkmark & \xmark & $25.14$ & $0.619$ & $0.516$   & $0.533$  &$1.143$   \\
                          & \xmark & \checkmark & \xmark & $29.81$     & $0.766$      & $0.513$        & $0.552$   & \colorbox{red!20}{1.895}     \\
                          & \checkmark & \xmark & \checkmark & $23.42$     & $0.582$      & $0.493$        & $0.505$   &$14.025$      \\
                          & \checkmark & \checkmark & \xmark & $25.39$     & $0.609$      & $0.482$        & $0.505$   &$17.147$      \\
                          & \checkmark & \xmark & \checkmark & $22.14$     & $0.533$      & $0.618$        & $0.567$   &$6.318$      \\ \hline
LerPlane                  & \checkmark & \checkmark & \xmark & $29.47$ & $0.766$ & $0.182$   & $0.292$  &$8.175$   \\
                          & \xmark & \checkmark & \xmark & $36.17$     & $0.900$      & \colorbox{blue!20}{0.139}     & $0.273$  & $20.875$ \\
                          & \checkmark & \xmark & \checkmark & \colorbox{red!20}{27.28}     & $0.710$      & $0.272$        & $0.355$  &$30.702$       \\
                          & \checkmark & \checkmark & \xmark & $32.70$     & $0.850$      & \colorbox{blue!20}{0.149}        & $0.214$   &$30.325$      \\
                          & \checkmark & \xmark & \checkmark & \colorbox{red!20}{26.11}     & $0.690$      & $0.296$        & $0.373$   &$28.511$      \\ \hline
LocalRF\(^{\dagger_2}\)
                  & \checkmark & \checkmark & \xmark & $29.02$ & $0.818$ & $0.177$   & $0.233$  &$3.926$   \\
                          & \xmark & \checkmark & \xmark & $35.07$ & $0.895$ & $0.172$   & \colorbox{blue!20}{0.235}  &$2.701$  \\
                          & \checkmark & \xmark & \checkmark & $22.54$     & $0.701$      & $0.318$        & $0.352$    &$6.424$     \\
                          & \checkmark & \checkmark & \xmark & $31.22$     & $0.841$      & $0.166$        & $0.209$   &$4.418$      \\
                          & \checkmark & \xmark & \checkmark & $19.21$     & $0.649$      & $0.394$        & $0.409$   &$5.411$      \\ \hline
HexPlane\(^{\dagger_1}\)
& \checkmark & \checkmark & \xmark & $31.50$ & \colorbox{red!20}{0.854} & $0.170$    & $0.233$   &\colorbox{red!20}{1.098}      \\
                          & \xmark & \checkmark & \xmark & \colorbox{red!20}{36.70}     & \colorbox{red!20}{0.916}    & $0.178$        & $0.251$  &$2.497$  \\
                          & \checkmark & \xmark & \checkmark & $27.13$     & $0.745$      & \colorbox{red!20}{0.258}        & $0.317$  & \colorbox{blue!20}{1.397}   \\
                          & \checkmark & \checkmark & \xmark & \colorbox{red!20}{33.04}     & \colorbox{red!20}{0.872}      & $0.162$        & \colorbox{red!20}{0.206}   &$1.154$      \\
                          & \checkmark & \xmark & \checkmark & $25.90$     & $0.710$      & $0.287$        & $0.359$   & \colorbox{blue!20}{1.516}      \\ \hline
Ours w/o Pose Optim.
                     & \checkmark & \checkmark & \xmark & \colorbox{blue!20}{31.91}     & \colorbox{blue!20}{0.875}    & \colorbox{red!20}{0.155}        & \colorbox{red!20}{0.201}   &$1.234$      \\
                          & \xmark & \checkmark & \xmark & \colorbox{blue!20}{37.03}     & \colorbox{blue!20}{0.917}    & \colorbox{red!20}{0.173}      & \colorbox{red!20}{0.242}  & \colorbox{red!20}{1.708}       \\
                          & \checkmark & \xmark & \checkmark & $27.21$     & \colorbox{red!20}{0.773}      & $0.242$        & \colorbox{red!20}{0.274}   &$1.533$      \\
                          & \checkmark & \checkmark & \xmark & \colorbox{blue!20}{33.71}     & \colorbox{blue!20}{0.889}      & \colorbox{red!20}{0.152}        & \colorbox{blue!20}{0.184}   & \colorbox{blue!20}{1.052}      \\
                          & \checkmark & \xmark & \checkmark & $25.65$     & \colorbox{red!20}{0.724}      & \colorbox{red!20}{0.279}        & \colorbox{red!20}{0.318}   &$1.752$      \\ \hline
Ours w/ Pose Optim.
& \checkmark & \checkmark & \xmark & \colorbox{red!20}{31.53}     & $0.851$      & \colorbox{blue!20}{0.139}        & \colorbox{blue!20}{0.196}    &\colorbox{blue!20}{1.052}     \\
                          & \xmark & \checkmark & \xmark & $35.25$     & $0.885$      & $0.180$        & $0.262$   &\colorbox{blue!20}{1.093}      \\
                          & \checkmark & \xmark & \checkmark & \colorbox{blue!20}{28.07}     & \colorbox{blue!20}{0.801}      & \colorbox{blue!20}{0.187}        & \colorbox{blue!20}{0.246}   & \colorbox{red!20}{1.453}      \\
                          & \checkmark & \checkmark & \xmark & $31.97$     & $0.816$     & $0.157$        & $0.216$    &\colorbox{red!20}{1.062}     \\
                          & \checkmark & \xmark & \checkmark & \colorbox{blue!20}{26.30}     & \colorbox{blue!20}{0.736}      & \colorbox{blue!20}{0.230}        & \colorbox{blue!20}{0.304}   & \colorbox{red!20}{1.703}      \\ \hline
\end{tabular}
}
\end{table}

\end{document}